# Classification of 24-hour movement behaviors from wrist-worn accelerometer data: from handcrafted features to deep learning techniques

Authors:

**Alireza Sameh[1], Mehrdad Rostami[2], Mourad Oussalah[2], Vahid Farrahi [1, 3,*]**

Affiliations:

1 Research Unit of Health Sciences and Technology, Faculty of Medicine, University of Oulu, Oulu, Finland

2 Centre of Machine Vision and Signal Analysis, Faculty of Information Technology and Electrical Engineering, University of Oulu, Oulu, Finland

3 Institute for Sport and Sport Science, TU Dortmund University, Dortmund, Germany

**\*Correspondence:**

**Vahid Farrahi, Institute for Sport and Sport Science, TU Dortmund University, Dortmund, Germany, vahid.farrahi@tu-dortmund.de**

**Abstract**

**Background:** Classical machine learning (ML) methods requiring handcrafted feature extraction have been widely used for predicting 24-hour movement behavior intensities from raw accelerometry signals. In contrast, deep learning (DL) approaches present an opportunity to automate feature extraction, which can be advantageous in handling the complexity of raw accelerometry signals. This study aimed to compare DL and classical ML algorithms for predicting movement behavior intensities from wrist-based raw accelerometry data into sleep, sedentary behavior, light-intensity (LPA) and moderate-to-vigorous physical activity (MVPA).

**Methods:** Open-access data from 151 adults wearing a wrist-worn accelerometer (Axivity-AX3) was used. Participants were randomly divided into training, validation, and test sets (121, 15, and 15 participants each). Raw acceleration signals were segmented into non-overlapping 10-second windows, and then a total of 104 handcrafted features (from time and frequency domains) were extracted. Four DL algorithms—Long Short-Term Memory (LSTM), Bidirectional Long Short-Term Memory (BiLSTM), Gated Recurrent Units (GRU), and One-Dimensional Convolutional Neural Network (1D-CNN)—were trained and tested using raw acceleration signals and with handcrafted features extracted from these signals to predict 24-hour movement behavior categories. Additionally, the handcrafted features were also used to train and test classical ML algorithms, namely Random Forest (RF), Support Vector Machine (SVM), Extreme Gradient Boosting (XGBoost), Logistic Regression (LR), Artificial Neural Network (ANN), and Decision Tree (DT) for classifying 24-hour movement behavior intensities.

**Results:** LSTM, BiLSTM, and GRU showed an overall accuracy of approximately 85% when trained with raw acceleration signals, and 1D-CNN an overall accuracy of approximately 80%. When trained on handcrafted features, the overall accuracy for both DL and classical ML algorithms ranged from 70% to 81%. Overall, there was a higher confusion in classification of MVPA and LPA, compared to sleep and sedentary categories. This confusion was lowest in RF, which achieved an accuracy of 93%, 76%, 66%, and 76% in predicting sleep, sedentary, LPA, and MVPA, respectively.

**Conclusion:** DL methods employing raw acceleration signals had only slightly better performance in predicting 24-hour movement behavior intensities, compared to when DL and classical ML were trained with handcrafted features. RF had overall better accuracy in classification of 24-hour movement behavior intensities.

# INTRODUCTION

In recent years, observational, population-based, and epidemiological studies have increasingly utilized wearable accelerometer-based activity monitors to measure movement behaviors, including physical activity, sedentary behavior, and sleep [1]. Unlike traditional self-reported methods—which are subjective and often prone to biases and inaccuracies, wearable accelerometers provide relatively more accurate data. However, the quantification of movement intensities from accelerometer outputs remains challenging [2]. There is still no universally accepted approach for characterizing movement intensities from accelerometer data [2].

In recent years, wrist-worn accelerometer-based activity monitors have gained popularity over other accelerometers due to their convenient size for daily use and their ease of wear compliance [3–5]. Traditionally, these wearable accelerometers were used for monitoring waking activity behaviors [6]. However, with the emergence of accelerometers capable of collecting raw accelerometer data, an increasing number of studies have shifted towards the use of accelerometer devices to measure movement behaviors around the clock. This has resulted in a paradigm shift in the literature towards 24-hour accelerometry, indicating that all movement behaviors within the 24-hour day including sleep, sedentary behaviors, light-intensity physical activity (LPA), and moderate-to-vigorous physical activity (MVPA) may be interdependently related to various health indicators [7]; this shift also reflects a growing interest in accurate prediction of 24-hour movement behavior intensities from raw acceleration data [8].

To date, many studies have used machine learning (ML) approaches to predict movement behavior intensities and categories from raw acceleration signals [8, 9]. Those studies utilizing classical ML algorithms such as random forest (RF) [10, 11], artificial neural networks (ANN) [11, 12], support vector machines (SVM) [11], and decision tree (DT) [11, 13] have typically reported acceptable accuracy in predicting intensities and categories. Such algorithms require handcrafted features, necessitating a process that is time-consuming, complex, and demands substantial domain expertise.

In practice, a considerable number of features from both the time and frequency domains (sometimes referred to as statistical features) can be extracted to classify movement behavior into various types, intensities, and categories using classical ML algorithms [14]. Currently, there is uncertainty regarding the optimal feature sets to extract to accurately predict 24-hour movement behavior; this may partly explain why most studies have continued to rely on cutoff-based techniques to translate acceleration signals into movement intensity categories [15].

The performance of ML techniques can vary significantly based on factors such as the quality and quantity of training data, feature selection and engineering, and model selection [16–18]. In recent years, deep learning (DL) algorithms have gained widespread use across many prediction tasks. One distinct characteristic of DL is its ability to automatically learn feature representations from raw data, eliminating the need for manually crafted feature extractors [19]. DL has already demonstrated excellence in numerous applications, such as signal processing tasks, due to its ability to automatically learn and extract discriminative features directly from raw signals [19]. Although DL methods have the potential to overcome the limitations associated with handcrafted feature extraction and can perform better than classical ML algorithms in predicting movement behaviors, their ability to predict 24-hour movement behaviors remains largely unclear [8].

DL algorithms operate based on a diverse range of concepts and principles. Human movement behavior data is, by nature, sequential [20]. Among the existing DL algorithms, some could be more suitable for analyzing sequential data since the algorithms use sequential information in their underlying principles. For example, Long Short-Term Memory (LSTM) networks are a frequently used algorithm for modeling sequential or time series data because of their capacity to capture long-term dependencies and temporal patterns. The unique gating mechanisms make them well-suited for understanding patterns within sequential dependencies and time series data [19, 21]. Few studies have used such DL algorithms for the classification of different dimensions of movement behavior from raw acceleration data [8, 9]. However, more research is needed to better understand the ability of DL algorithms to predict 24-hour movement behaviors into sleep, sedentary, LPA, and MVPA [9, 22].

The main contribution of this study lies in its comprehensive comparative analysis of classical ML and DL algorithms for classifying 24-hour movement behavior intensities into four categories (sleep, sedentary, LPA, and MVPA). While previous studies have explored various ML and DL methods for predicting movement intensity from accelerometer data, uncertainty remains regarding which approach performs better. This study evaluates four DL models—LSTM, Bidirectional LSTM (BiLSTM), Gated Recurrent Unit (GRU), and one-dimensional Convolutional Neural Network (1D-CNN)—which can utilize raw acceleration data or handcrafted features as input. These models are compared with classical ML algorithms, Extreme Gradient Boosting (XGBoost), RF, ANN, SVM, DT, and Logistic Regression (LR) which rely on handcrafted features as inputs to predict 24-hour movement behavior intensities.

## METHODS

The study procedure is visually depicted in Fig. 1. The scenarios were conducted to compare DL algorithms with classical ML algorithms in predicting 24-hour movement behavior intensities. First, we applied DL algorithms directly to raw acceleration signals for classification of the 24-hour movement behaviors. Following the existing literature [8, 13, 16, 23, 24], we also created a comprehensive list of time domain and frequency domain feature sets from raw acceleration signals. These handcrafted features were used as input to both DL and classical ML algorithms for the classification of 24-hour movement behavior intensities. When trained with handcrafted features, feature selection could be an important step. This is because input features can directly influence the predictive performance of DL and ML algorithm [14], but the most important features for a given prediction task are typically not known a priori. To investigate whether the predictive performance of DL and ML improves when trained on the most important features, we identified top subsets of handcrafted features using the widely used feature selection method [25] and used them as inputs for the DL and ML algorithms.

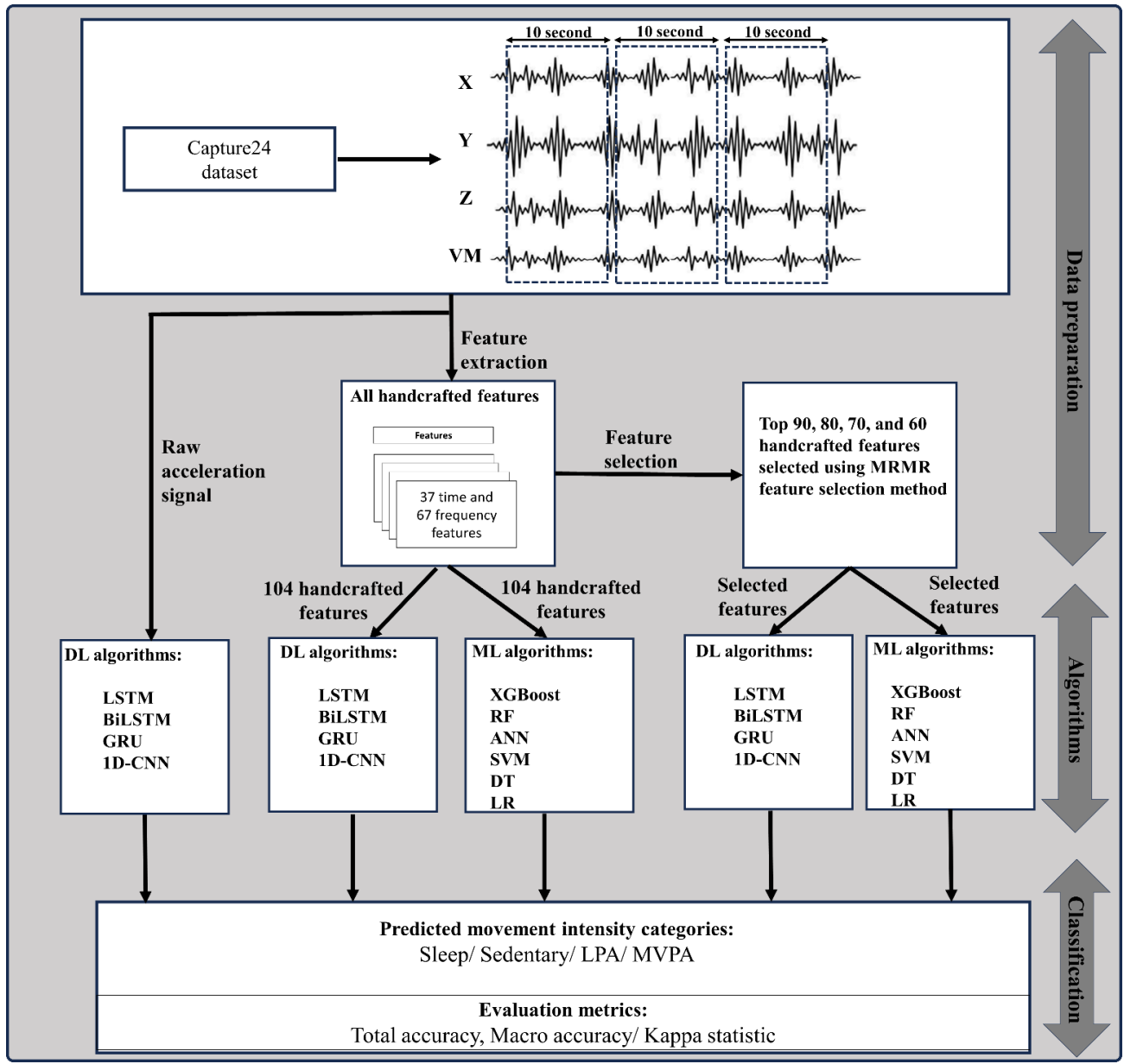

**Figure 1: Study flow, presenting the experiments performed with DL and ML algorithms. Abbreviations: 1D-CNN (One Dimensional Convolutional Neural Network), ANN (Artificial Neural Network), BiLSTM (Bidirectional Long Short-Term Memory), DL (Deep learning), DT (Decision Tree), GRU (Gated Recurrent Unit), LPA (Light Physical Activity), LR (Logistic Regression), LSTM (Long Short-Term Memory), ML (Machine learning), MVPA (Moderate to Vigorous Physical Activity), RF (Random Forest), SVM (Support Vector Machine), and XGBoost (eXtreme Gradient Boosting).**

## Dataset

This study used the publicly available Capture-24 dataset, which is an accelerometer validation study encompassing data from 151 adults aged from 18 to 91 years [23, 26]. Participants were asked to continuously wear an Axivity AX3 wrist-worn accelerometer, which captures raw signals at a frequency of 100 Hz, for a duration of 24 hours. Participants were also instructed to wear a Vicon Autographer wearable camera on their chest while awake and to maintain a time-use diary to record their sleep time. The wearable camera was programmed to capture images at regular 30-second intervals, and these images were then used to identify activity types. A more description of the dataset and the recruitment process is presented elsewhere [23, 26].

### *Annotations and 24-hour movement behavior*

Trained researchers annotated accelerometer data based on camera images and hierarchical labels derived from the Compendium of Physical Activities and [27]. Briefly, sleep periods were identified and marked based on the information recorded in the diary. To establish the 24-hour movement behavior categories, the activity types were mapped to sleep, sedentary behavior, LPA, and MVPA. This categorization was based on the body posture and activity energy expenditure values provided in the Compendium. Further details about the annotation process can be found elsewhere [23, 26].

### Data preparation

### *Signal segmentation*

Both to utilize raw acceleration signals directly and to extract handcrafted features, the signals were segmented into fixed-length, non-overlapping windows. Selecting a large window size might result in missing brisk and sporadic activities [24], whereas a shorter window size could pose challenges in identifying and capturing the meaningful signal features. Since the dataset used in this study includes a wide range of free-living activities [8], the signals were segmented into 10-second windows. This window length has been widely used in the existing literature and is considered sufficient for both DL and classical ML approaches to classify movement behavior intensity and categories [9, 24].

### *Handcrafted feature extraction*

Following prior research [8, 13, 16, 23, 28], we extracted a comprehensive list of features from all acceleration axes (x, y, z) as well as the vector magnitude ($VM = \sqrt{x^2 + y^2 + z^2}$) [13]. For each 10-second window, a total number of 104 features including 37 time domain features and 67 frequency domain features were computed [8, 23]. These features have all been previously used for movement behavior type and intensity classification from raw acceleration signals in the existing literature, and they have been considered appropriate for predicting sleep, sedentary behaviors, LPA, and MVPA from wrist-worn raw accelerometry data. The list of extracted features is shown in Table 1. To ensure the informativeness of these features for predicting 24-hour movement behavior intensities, we calculated mutual information metric for the extracted features. All features were found to contain relevant information (see **Supplementary Materials Fig. S1**). Mutual information is a widely used metric to assess whether features contain relevant information for a prediction task [25].

**Table 1: List of extracted features (handcrafted features) from raw acceleration signal**

| Domain | Features | Axis |
|---|---|---|
| **Time domain** | Mean, Standard deviation, Maximum, Minimum | x, y, z, VM |
| | 25th, 50th, 75th percentile | x, y, z, VM |
| | Correlation (x,y), correlation (x, z), correlation (y, z) | x, y, z |
| | Average of $\tan^{-1}$(y, x) value | x, y |
| | Standard deviation of yaw | x, y |
| | Average of $\tan^{-1}$(x, z) values (pitch) | x, z |
| | Standard deviation of pitch | x, z |
| | Average of $\tan^{-1}$(y, z) values (roll) | y, z |
| | Standard deviation of roll | y, z |
| **Frequency domain** | Total power | VM |
| | Dominant frequency | VM |
| | Power of the dominant frequency | VM |
| | Second dominant frequency | VM |
| | Power of second dominant frequency | VM |
| | Dominant frequency between 0.6 and 2.5 Hz | VM |
| | Power corresponding | VM |

| Power at 1Hz, 2Hz, 3Hz, 4Hz, 6Hz, 7Hz, 8Hz, 9Hz, 10Hz, 11Hz, 12Hz, 13Hz, 14Hz, 15Hz | x, y, z, VM |

## Deep learning algorithms

This study employed a total of four DL algorithms recognized for their ability to handle sequential and long-term dependency datasets. Here, we provide a brief overview of the principles and theories underlying each of these algorithms. Detailed descriptions of LSTM [29], BiLSTM [30], GRU [31], and CNN [32] can be found elsewhere.

### Long short-term memory (LSTM)

LSTM networks are a specialized type of Recurrent neural networks (RNNs) designed to learn long-term dependencies within variable-length sequential data. Traditional RNNs often struggle to capture long-term dependencies effectively [29]. LSTM can learn dependencies across a substantial number of discrete time steps. This property makes them an appropriate choice to be coupled with time series data, where temporal patterns and dependencies are important to consider.

### Bidirectional long short-term memory (BiLSTM)

The BiLSTM algorithm is characterized by its ability to process sequential data bidirectionally. It overcomes the limitations of the traditional LSTM algorithm by connecting the output layer to two hidden LSTM layers [19] enabling information flow in both forward and backward directions. This unique design allows the BiLSTM algorithm to take full advantage of historical information by utilizing a forget gate to selectively retain or discard relevant information [30]. By considering information from both the past and the future, BiLSTM enables a comprehensive understanding of the sequential data. This bidirectional approach enhances its ability to capture long-term dependency patterns, making it well-suited for modeling complex sequential relationships [33].

### Gated recurrent units (GRU)

The GRU is a simplified version of the LSTM model. Unlike LSTM, GRU does not have a separate memory cell to store information. Instead, it can only control and modify the information within the unit itself [19]. This algorithm consists of two gates: the "update gate," which includes input and forget gates, and the "reset gate". This structure lets GRU capture dependencies from big data sequences [31].

### *Convolutional neural networks (CNNs)*

Convolutional neural networks (CNNs) typically comprise input layers, hidden layers, and an output layer, which can be a fully connected layer. The hidden layers within CNNs encompass convolutional layers, rectified linear units, and pooling layers, each serving specific functions [19]. Traditional two-dimensional convolutional neural networks (2D CNNs) are commonly employed for pattern recognition and classification tasks, particularly with image data. However, an alternative approach using 1D-CNNs has the capacity to process signals as well. This algorithm architecture is specifically designed to be applied to one-dimensional data, such as signals and time series data, and utilizes convolutional operations for processing [19, 34].

### Classical machine learning algorithms

The classical ML algorithms considered in this study include XGBoost, RF, ANN, SVM, DT, and LR. These algorithms have all previously been employed for the classification of movement behavior intensity and category. Brief descriptions of the algorithms are provided below [10–13, 24].

### *Extreme gradient boosting (XGBoost)*

XGBoost is an advanced version of the gradient-boosting classifier, focusing on speed and performance. It uses regularized learning to prevent overfitting, which can occur when the model becomes too specialized to the training data. The objective function of XGBoost combines a convex loss function with a penalty term to strike a balance between minimizing training loss and controlling model complexity, ultimately leading to improved predictions on new data [35].

***Random forest (RF)***

RF is a collection of DTs used for classification tasks. In a DT algorithm, a tree structure is employed to represent the data, with each leaf node corresponding to a class label and internal nodes representing attributes. The challenge lies in selecting the appropriate attribute for each node at every level, which is typically achieved using metrics like information gain and Gini index. RF constructs each constituent tree by using a bootstrap sample from the training data. In cases of highly imbalanced data, there is a notable chance that a bootstrap sample might contain only a few instances from the minority class. This can lead to the development of individual trees with suboptimal performance in predicting the minority class [36].

***Artificial neural networks (ANNs)***

ANNs are mathematical models inspired by the biological nerve systems in the human brain. Multilayer Perceptron (MLP) is a type of ANN that manages non-linear models by adapting weight factors and biases through a gradient descent process aimed at minimizing the target error. The training process in ANN involves three main steps: 1) forward propagation, where input data passes through different layers to generate outputs; 2) computation of the error by comparing predicted and actual outputs; 3) backward propagation, which adjusts the synaptic weights to reduce the error at each step, bringing the ANN model closer to the desired output [37].

***Support vector machine (SVM)***

SVM operates as a classification technique, creating decision boundaries (hyperplanes) in a multi-dimensional space to distinguish instances with distinct class labels. The primary goal of this algorithm is to identify a hyperplane with the maximum margin, representing the greatest distance between the hyperplane and data points from classes. By maximizing the margin, SVM ensures enhanced confidence in classifying future data points. SVM can utilize kernel functions to transform data into higher-dimensional spaces, thereby improving the distinguishability of the data [38].

*Decision tree (DT)*

This algorithm creates a tree-shaped model where internal nodes signify features, branches indicate decisions based on feature values, and leaf nodes offer predictions for the target variable. Predictions with decision trees entail moving through the tree from the root node to a leaf node, choosing branches based on input feature values. The prediction at the leaf node represents the anticipated value of the target variable [39].

*Logistic regression (LR)*

LR is a type of model used for classification tasks where the goal is to predict the probability of an instance belonging to a specific class. In this approach, we have a categorical response variable $Y$ and a set of independent variables $X = \{X_1, X_2, \dots, X_n\}$ that are used to make predictions. The core idea of logistic regression is to model the relationship between the independent variables and the response variable as a linear function, represented by coefficients $W = \{W_0, W_1, W_2, \dots, W_n\}$. The training process in logistic regression involves adjusting the coefficients $W$ to maximize the cross-entropy between the model's predicted probabilities and the true class labels. This optimization process aims to find the best-fitting model that can accurately classify instances based on given independent variables [40].

**Experiments**

*Deep learning with raw acceleration signal*

A key advantage of DL algorithms over classical ML algorithms is their ability to automatically learn and extract feature representations from raw data. Segmented 10-second raw acceleration signals were directly used as input to LSTM, BiLSTM, GRU, and 1D-CNN for classifying movement behavior intensities as sleep, sedentary, LPA, and MVPA. The 10-second raw acceleration signals were used in their original form (x, y, z, VM) as input; this means the input size was a 4 ×1000 matrix. Detailed information about each DL architecture can be found in Fig. 2.

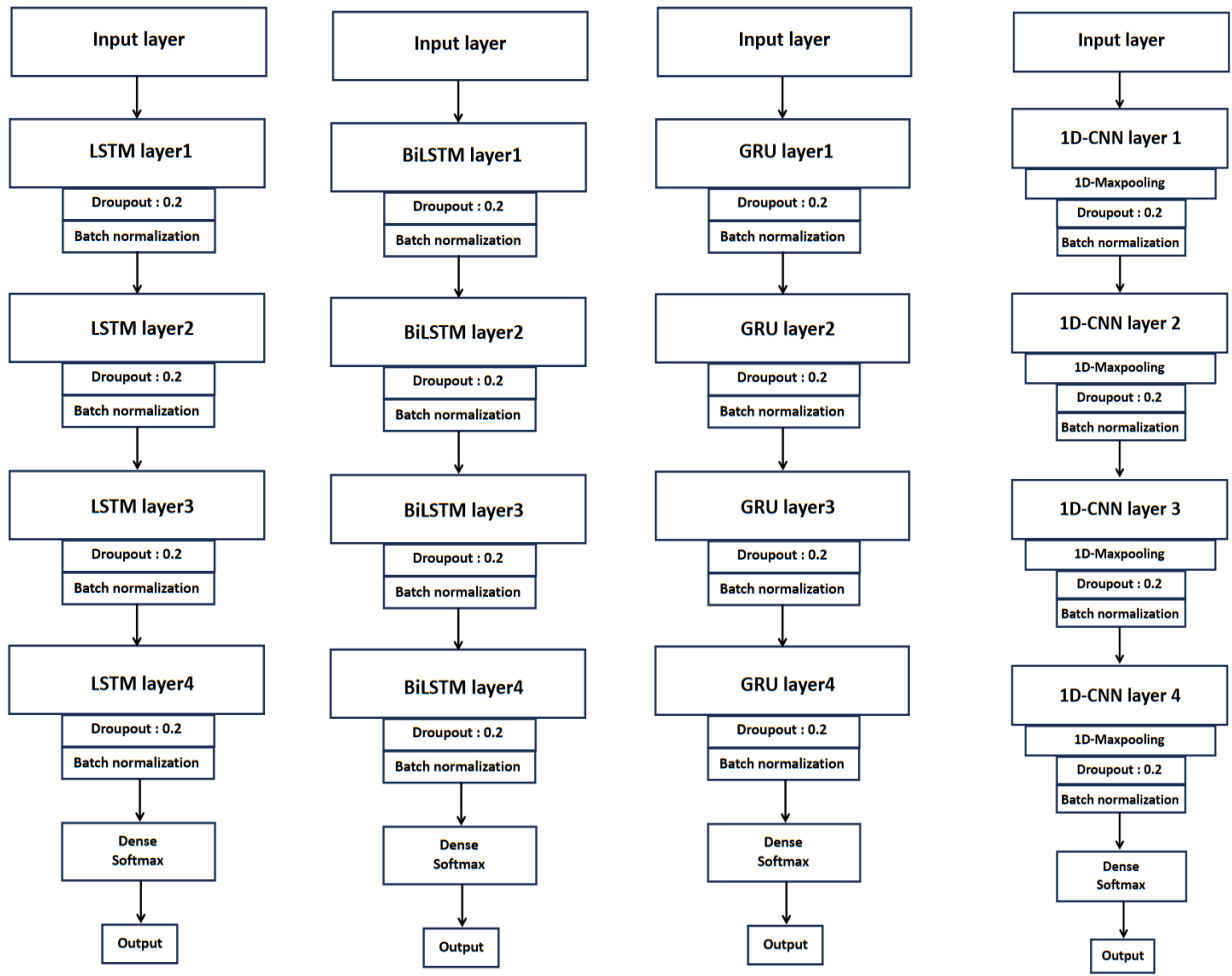

**Figure 2: The architecture used for training DL algorithms. Abbreviations: 1D-CNN (One Dimensional Convolutional Neural Network), ANN (Artificial Neural Network), BiLSTM (Bidirectional Long Short-Term Memory), GRU (Gated Recurrent Unit), LSTM (Long Short-Term Memory).**

### *Deep learning and classical machine learning with all 104 handcrafted features*

In addition to raw acceleration signal, handcrafted features can also be used as input to DL algorithms. All the handcrafted features (in total 104 features) extracted from the 10-second window intervals, were used as inputs for LSTM, BiLSTM, GRU, and 1D-CNN algorithms. The architecture employed for the DL algorithms were similar to when these algorithms used raw acceleration signal as inputs (see Fig. 2).

Classical ML algorithms require prior feature extraction before training, often relying on domain knowledge to define relevant features. The same set of 104 handcrafted features was also used as inputs to train six classical ML algorithms for predicting 24-hour movement behaviors. These ML algorithms were XGBoost, RF, ANN, SVM, DT, and LR.

### *Deep learning and classical machine learning with selected handcrafted features*

When utilizing handcrafted features, the set of input feature may influence the predictive performance of ML and DL algorithms [16]. We conducted a feature selection method to determine whether and how the results of experiments using handcrafted features were affected by the choice of input features. We employed the widely used minimum redundancy-maximum relevance (MRMR) feature selection method to select the most informative features and re-trained the models [25]. MRMR is a well-established technique frequently utilized in ML for selecting appropriate features for various classification tasks. It has been demonstrated to be effective in choosing feature sets with minimal redundant information [25]. Using the MRMR method, we selected the top 90, 80, 70, and 60 most relevant and non-redundant features, which were then used as inputs to both DL and ML algorithms.

### **Structure of deep learning methods and hyperparameter tuning for classical machine learning**

The structures for DL methods were experimentally selected. These experiments included increasing the number of layers from 1 to 5 with batch normalization and dropout after each layer with early stopping. Overfitting is a common issue in machine learning, especially for neural networks due to their flexible nature. Typically, early stopping and dropout regularization techniques are utilized to address the problem of overfitting in DL algorithms [33]. Dropout temporarily removes neurons and connections during training based on a specified probability. This prevents the network from overly adapting to the training data, as well as helping to mitigate overfitting and improving generalization on unseen data. Early stopping was done based on the validation loss [33].

We also carefully selected and fine-tuned the parameters for the XGBoost, RF, ANN, SVM, DT, and LR algorithms using grid search to optimize total accuracy within the range specified by previous studies [14, 26, 41]. The detailed parameters used for training these algorithms are provided in Table 2.

**Table 2: Hyperparameters used for training classical ML algorithms.**

| Algorithm | Hyperparameters |
|---|---|
| **Random forest (RF)** | Each random forest contained 500 trees, and the size of the random subset of features at each split was the square root of the total number of features. Sampling with replacement was done for all classes |
| **Artificial neural networks (ANN)** | Each network contained a single hidden layer of 15 nodes and was trained for a maximum of 200 iterations. |
| **Support vector machines (SVM)** | The kernel type was radial basis function 'rbf', and the regularization parameter was set to 200. |
| **Logistic regression (LR)** | The maximum iteration of was set to 1000 and solver was selected 'lbfgs' (Limited-memory Broyden–Fletcher–Goldfarb–Shanno). the random state was set to 17 |
| **Decision tree (DT)** | The minimum number of samples required to split an internal node was set to 100, and the minimum number of samples required to be at a leaf node was set to 50 |
| **Extreme gradient boosting (XGBoost)** | The maximum tree depth was set to 20, maximum number of estimators 50, lambda for L2 regularization |

## Training, validation, and test sets

We performed a random split of the entire dataset into training, validation, and testing sets. The training set consisted of data from 121 participants, which accounted for 80% of the dataset. The testing and validation set each included data from 15 participants, representing 10% of the dataset. The dataset was divided using a participant-based approach following the widely used leave-one-participant-out procedure [8, 42] to ensure realistic evaluation and generalizability of the results. We trained the algorithms using the training set and fine-tuned their hyperparameters using the validation set. Finally, we evaluated the performance of the trained algorithms using the unseen data from the test set.

## Performance evaluation and sensitivity analysis

To evaluate the performance of each method, confusion matrices were employed to assess classification and misclassification rates, as well as overall classification accuracy on test sets. We also calculated macro accuracy and Kappa statistics (K) which are essential metrics for evaluating classification models, particularly in multiclass and imbalanced data settings [43]. Macro accuracy calculates the average accuracy scores for each class, ensuring each class contributes equally, irrespective of size. Kappa statistic measure the agreement between observed and predicted classifications, adjusting for chance agreement, providing a robust measure of model reliability beyond simple accuracy [43].

To ensure that the algorithms were robust, a sensitivity analysis was conducted to ensure the robustness of all algorithms to variations in the training set. This was achieved by repeating the training and testing processes with the same parameters while exchanging participant data initially assigned to the test sets with data from 15 randomly selected participants from the training set. All the experimental setups were implemented in Python using the scikit-learn and TensorFlow libraries.

## RESULTS

### Classification performance of deep learning with raw acceleration signals

The confusion matrices, depicting the performance of DL algorithms with raw acceleration signals, are presented in Fig. 3. LSTM, BiLSTM, and GRU algorithms demonstrated relatively better performance than 1D-CNN, achieving a total accuracy of approximately 85%. The 1D-CNN algorithms had a total accuracy of 80%. Overall, DL algorithms performed better in the classification of sleep and sedentary classes (range: 94%–96.5% for sleep and 75%–86% for sedentary). However, they had relatively lower accuracy when classifying LPA (63%–76%) and MVPA (47%–60%). GRU demonstrated the highest macro accuracy at 78.2%. This was followed by LSTM and BiLSTM, which achieved macro accuracies in a similar range of 75%-76. The 1D-CNN, with a macro accuracy of 72.6%, exhibited the lowest performance among the models. LSTM, BiLSTM, and GRU all demonstrated equal kappa statistics of 0.76. In contrast, the 1D-CNN recorded a lower kappa statistic of 0.69.

**Long-short term memory (LSTM)**

| Class | Sleep | Sedentary | LPA | MVPA | |
|---|---|---|---|---|---|
| Sleep | [23861] 94.8 % | [1171] 4.7 % | [130] 0.5 % | [1] 0.0 % | Total accuracy: 85% |
| Sedentary | [2683] 8.4 % | [26644] 83.7 % | [2380] 7.5 % | [119] 0.4 % | Macro accuracy: 75.4 % |
| LPA | [199] 1.7 % | [2448] 20.6 % | [9082] 76.3 % | [176] 1.5 % | Kappa statistics, K: 0.76 |
| MVPA | [1] 0.0 % | [120] 4.4 % | [1309] 48.5 % | [1268] 47.0 % | |

**Bidirectional Long-short term memory (BiLSTM)**

| Class | Sleep | Sedentary | LPA | MVPA | |
|---|---|---|---|---|---|
| Sleep | [23666] 94.1 % | [1381] 5.5 % | [116] 0.5 % | [0] 0.0 % | Total accuracy: 85% |
| Sedentary | [2435] 7.7 % | [27390] 86.1 % | [1937] 6.1 % | [64] 0.2 % | Macro accuracy: 75.9 % |
| LPA | [194] 1.6 % | [2901] 24.4 % | [8366] 70.3 % | [444] 3.7 % | Kappa statistics, K: 0.76 |
| MVPA | [2] 0.1 % | [124] 4.6 % | [1134] 42.0 % | [1438] 53.3 % | |

**Gated Recurrent Units (GRU)**

| Class | Sleep | Sedentary | LPA | MVPA | |
|---|---|---|---|---|---|
| Sleep | [23928] 95.1 % | [1025] 4.1 % | [207] 0.8 % | [3] 0.0 % | Total accuracy: 84.91 % |
| Sedentary | [2787] 8.8 % | [26318] 82.7 % | [2637] 8.3 % | [84] 0.3 % | Macro accuracy: 78.2 % |
| LPA | [225] 1.9 % | [2068] 17.4 % | [8924] 75.0 % | [688] 5.8 % | Kappa statistics, K: 0.76 |
| MVPA | [0] 0.0 % | [83] 3.1 % | [995] 36.9 % | [1620] 60.0 % | |

**One-dimensional Convolutional Neural Network (1D-CNN)**

| Class | Sleep | Sedentary | LPA | MVPA | |
|---|---|---|---|---|---|
| Sleep | [24293] 96.5 % | [771] 3.1 % | [98] 0.4 % | [1] 0.0 % | Total accuracy: 80 % |
| Sedentary | [5951] 18.7 % | [23975] 75.3 % | [1869] 5.9 % | [31] 0.1 % | Macro accuracy: 72.6 % |
| LPA | [684] 5.7 % | [3179] 26.7 % | [7527] 63.2 % | [515] 4.3 % | Kappa statistics, K: 0.69 |
| MVPA | [15] 0.6 % | [163] 6.6 % | [1024] 38.0 % | [1496] 55.4 % | |

**Figure 3: Confusion matrix and evaluation metrics of DL algorithms on the test set (n = 15 participants) when trained on raw acceleration data. Abbreviations: Light physical activity (LPA), Moderate to vigorous physical activity (MVPA).**

**Classification performance of deep learning with handcrafted features**

Fig. 4 presents the confusion matrices of the performance of DL algorithms when trained with handcrafted features. Overall, the total accuracy of DL algorithms when trained with handcrafted features was lower than when the same algorithms were trained with raw acceleration signals, achieving a total accuracy of approximately 79%–80%. LSTM, BiLSTM, GRU, and 1D CNN when trained with handcrafted features were found to be less accurate in the classification of sleep (88%–91%), sedentary (80%–83%), and LPA (60%–67%), compared to when these algorithms were trained with raw acceleration signal. GRU achieved the highest macro accuracy of 71.2%, paired with a kappa statistic of 0.69. LSTM closely followed with a macro accuracy of 70.9% and a slightly lower kappa of 0.68. BiLSTM exhibited a macro accuracy of 69.8%, slightly behind the GRU and LSTM models, but matched GRU with a kappa of 0.69. 1D-CNN demonstrated the highest macro accuracy of 74.8% among all algorithms and achieved the best kappa statistic of 0.7 in this experiment.

| Long-short term memory (LSTM) | | | | |
|---|---|---|---|---|
| Class | Sleep | Sedentary | LPA | MVPA | |
| Sleep | [25287] 88.2% | [3260] 11.4% | [112] 0.4% | [3] 0.0% | Total accuracy: 79.9% |
| Sedentary | [3570] 9.8% | [30456] 83.4% | [2423] 6.6% | [55] 0.2% | Macro accuracy: 70.9% |
| LPA | [204] 1.5% | [4703] 33.4% | [8525] 60.6% | [630] 4.5% | Kappa statistics, K: 0.68 |
| MVPA | [54] 1.6% | [215] 6.4% | [1368] 40.8% | [1734] 51.4% | |

| Bidirectional Long-short term memory (BiLSTM) | | | | |
|---|---|---|---|---|
| Class | Sleep | Sedentary | LPA | MVPA | |
| Sleep | [25986] 90.7% | [2495] 8.7% | [176] 0.6% | [5] 0.0% | Total accuracy: 80.2% |
| Sedentary | [3316] 9.1% | [29424] 80.6% | [3730] 10.2% | [34] 0.1% | Macro accuracy: 69.8% |
| LPA | [178] 1.3% | [3824] 27.2% | [9510] 67.6% | [550] 3.9% | Kappa statistics, K: 0.69 |
| MVPA | [3] 0.1% | [172] 5.1% | [1835] 54.4% | [1361] 40.4% | |

| Gated Recurrent Units (GRU) | | | | |
|---|---|---|---|---|
| Class | Sleep | Sedentary | LPA | MVPA | |
| Sleep | [25322] 88.3% | [3194] 11.1% | [144] 0.5% | [2] 0.0% | Total accuracy: 79.9% |
| Sedentary | [3336] 9.1% | [29817] 81.7% | [3277] 9.0% | [74] 0.2% | Macro accuracy: 71.2% |
| LPA | [180] 1.3% | [4050] 28.8% | [9244] 65.7% | [588] 4.2% | Kappa statistics, K: 0.69 |
| MVPA | [2] 0.1% | [176] 5.2% | [1537] 45.6% | [1656] 49.1% | |

| One-dimensional Convolutional Neural Network (1D-CNN) | | | | |
|---|---|---|---|---|
| Class | Sleep | Sedentary | LPA | MVPA | |
| Sleep | [26123] 91.1% | [2397] 8.4% | [135] 0.5% | [7] 0.0% | Total accuracy: 80.6% |
| Sedentary | [3533] 9.7% | [29554] 81.0% | [3279] 9.0% | 138 0.4% | Macro accuracy: 74.8% |
| LPA | [227] 1.6% | [4095] 29.1% | [8758] 62.3% | [982] 7.0% | Kappa statistics, K: 0.70 |
| MVPA | [3] 0.1% | [166] 4.9% | [1020] 30.3% | [2182] 64.7% | |

**Figure 4 : Confusion matrix and evaluation metrics of DL algorithms on the test set (n = 15 participants) when trained with 104 handcrafted features extracted from time and frequency signal domain. Abbreviations: Light physical activity (LPA), Moderate to vigorous physical activity (MVPA).**

**Classification performance of classical machine learning with handcrafted features**

Fig. 5 presents the performance of classical ML algorithms, including XGBoost, RF, ANN, SVM, DT, and LR. Among classical ML algorithms, XGBoost achieved the highest classification total accuracy of 81.1%, followed by RF and ANN (both 80%). SVM, DT, and LR had relatively lower performance with a total accuracy of 78%, 76%, and 70%, respectively. Among these algorithms, RF had less confusion in classification of LPA and MVPA. RF algorithm achieved the highest macro accuracy of 77.8% along with a kappa statistic of 0.7. Similarly, the XGBoost algorithm exhibited a competitive macro accuracy of 72.2% and matched RF with a kappa of 0.7. ANN achieved a macro accuracy of 71.4% with a kappa of 0.7. SVM showed a slightly lower macro accuracy of 67.7% and a corresponding kappa of 0.67. DT exhibited a macro accuracy of 68.3% but had a lower kappa of 0.6. LR algorithm reported the lowest macro accuracy of 63.8% and the lowest kappa statistic of 0.54.

**eXtreme Gradient Boosting (XGBoost)**

| Class | Sleep | Sedentary | LPA | MVPA | |
|---|---|---|---|---|---|
| Sleep | [26306] 91.8 % | [2215] 7.7 % | [141] 0.5 % | [0] 0.0 % | Total accuracy: 81.1 % |
| Sedentary | [3739] 10.2 % | [30157] 82.6 % | [2578] 7.1 % | [30] 0.1 % | Macro accuracy: 72.2 % |
| LPA | [196] 1.4 % | [4419] 31.4 % | [8826] 62.8 % | [621] 4.4 % | Kappa statistics, K: 0.7 |
| MVPA | [3] 0.1 % | [220] 6.5 % | [1409] 41.8 % | [1739] 51.6 % | |

**Random Forest (RF)**

| Class | Sleep | Sedentary | LPA | MVPA | |
|---|---|---|---|---|---|
| Sleep | [26575] 92.7 % | [1800] 6.3 % | [265] 0.9 % | [22] 0.1 % | Total accuracy: 80.2 % |
| Sedentary | [3964] 10.9 % | [27756] 76.0 % | [4611] 12.6 % | [173] 0.5 % | Macro accuracy: 77.8 % |
| LPA | [202] 1.4 % | [3142] 22.3 % | [9332] 66.4 % | [1386] 9.9 % | Kappa statistics, K: 0.7 |
| MVPA | [0] 0.0 % | [122] 3.6 % | [686] 20.4 % | [2563] 76.0 % | |

**Artificial Neural Network (ANN)**

| Class | Sleep | Sedentary | LPA | MVPA | |
|---|---|---|---|---|---|
| Sleep | [30274] 92.2 % | [2385] 7.3 % | [157] 0.5 % | [3] 0.0 % | Total accuracy: 80 % |
| Sedentary | [3675] 10.8 % | [28010] 82.0 % | [2395] 7.0 % | [68] 0.2 % | Macro accuracy: 71.4 % |
| LPA | [111] 0.7 % | [4755] 32.0 % | [9218] 62.0 % | [779] 5.2 % | Kappa statistics, K: 0.7 |
| MVPA | [26] 0.4 % | [936] 14.6 % | [2296] 35.7 % | [3170] 49.3 % | |

**Support Vector Machine (SVM)**

| Class | Sleep | Sedentary | LPA | MVPA | |
|---|---|---|---|---|---|
| Sleep | [26857] 93.7 % | [1695] 5.9 % | [109] 0.4 % | [1] 0.0 % | Total accuracy: 78.7 % |
| Sedentary | [5939] 16.3 % | [28289] 77.5 % | [2257] 6.2 % | [19] 0.1 % | Macro accuracy: 67.7 % |
| LPA | [570] 4.1 % | [4568] 32.5 % | [8524] 60.6 % | [400] 2.8 % | Kappa statistics, K: 0.67 |
| MVPA | [7] 0.2 % | [271] 8.0 % | [1783] 52.9 % | [1310] 38.9 % | |

**Decision Tree (DT)**

| Class | Sleep | Sedentary | LPA | MVPA | |
|---|---|---|---|---|---|
| Sleep | [24249] 84.6 % | [4179] 14.6 % | [226] 0.8 % | [8] 0.0 % | Total accuracy: 76.3 % |
| Sedentary | [3953] 10.8 % | [29032] 79.5 % | [3448] 9.4 % | [71] 0.2 % | Macro accuracy: 68.3 % |
| LPA | [244] 1.7 % | [4874] 34.7 % | [8364] 59.5 % | [580] 4.1 % | Kappa statistics, K: 0.63 |
| MVPA | [34] 1.0 % | [256] 7.6 % | [1678] 49.8 % | [1403] 41.6 % | |

**Logistic Regression (LR)**

| Class | Sleep | Sedentary | LPA | MVPA | |
|---|---|---|---|---|---|
| Sleep | [25718] 89.7 % | [2851] 9.9 % | [93] 0.3 % | [2] 0.0 % | Total accuracy: 70.4 % |
| Sedentary | [10943] 30.0 % | [23545] 64.5 % | [1994] 5.5 % | [22] 0.1 % | Macro accuracy: 63.8 % |
| LPA | [1382] 9.8 % | [4895] 34.8 % | [7189] 51.1 % | [596] 4.2 % | Kappa statistics, K: 0.54 |
| MVPA | [49] 1.5 % | [403] 12.0 % | [1240] 36.8 % | [1679] 49.8 % | |

**Figure 5: Confusion matrix and evaluation metrics of ML algorithms on the test set (n = 15 participants) when trained with 104 handcrafted features extracted from time and frequency signal domain. Abbreviations: Light physical activity (LPA), Moderate to vigorous physical activity (MVPA).**

## Classification performance of deep learning and classical machine learning with selected handcrafted features

The results of the feature selection, which involved training algorithms with the top 60, 70, 80, and 90 handcrafted features ranked by the MRMR feature selection method, are presented in Table 3. Results show that reducing the number of features using the MRMR method had a minimal impact on the performance metrics of both DL and classical ML algorithms. For DL algorithms, total accuracy, macro accuracy, and Kappa statistics remained stable despite feature reduction, with variations of only 1–2%. In classical ML, similarly total accuracy, macro accuracy, and Kappa statistics remained stable despite feature reduction, with variations of only 1-4%.

**Table 3: Performance metrics when the algorithms were trained with features selected with all features, and 90, 80, 70, and 60 features selected with Minimum-Redundancy-Maximum-Relevance (MRMR) feature selection (all numbers were rounded to the nearest value and the highest performing values are shown in bold)**

| Algorithm | Performance metric | Number of features | | | | |
| --- | --- | --- | --- | --- | --- | --- |
| | | 104 (All features) | 90 | 80 | 70 | 60 |
| ***Deep learning algorithms*** | | | | | | |
| Long short-term memory (LSTM) | Total accuracy | **80 %** | 79 % | **80 %** | 79 % | 79 % |
| | Macro accuracy | **70 %** | 70 % | **70 %** | 70 % | 70 % |
| | Kappa statistics, K | **0.68** | 0.68 | **0.68** | 0.68 | 0.68 |
| Bi-directional long short-term memory (BiLSTM) | Total accuracy | **80 %** | 79 % | 79 % | 79 % | 79 % |
| | Macro accuracy | **69 %** | 69 % | 69 % | 69 % | 69 % |
| | Kappa statistics, K | **0.69** | 0.69 | 0.69 | 0.69 | 0.69 |
| Gated recurrent unit (GRU) | Total accuracy | **80 %** | 79 % | 79 % | 79 % | 79 % |
| | Macro accuracy | **71 %** | 71 % | 71 % | 70 % | 71 % |
| | Kappa statistics, K | **0.69** | 0.69 | 0.69 | 0.69 | 0.69 |
| One- dimensional convolutional neural network (1D-CNN) | Total accuracy | **80 %** | 79 % | 79 % | 79 % | 78 % |
| | Macro accuracy | **74 %** | 74 % | 74 % | 74 % | 74 % |
| | Kappa statistics, K | **0.7** | 0.7 | 0.7 | 0.7 | 0.7 |
| ***Classical ML algorithms*** | | | | | | |
| eXtreme Gradient Boosting (XGBoost) | Total accuracy | **81 %** | **81 %** | 80 % | 79 % | 79 % |
| | Macro accuracy | **72 %** | **72 %** | 72 % | 72 % | 72 % |
| | Kappa statistics, K | **0.7** | **0.7** | 0.7 | 0.7 | 0.7 |
| Random forest (RF) | Total accuracy | **80 %** | **80 %** | 80 % | 79 % | 79 % |
| | Macro accuracy | **78 %** | **78 %** | 77 % | 77 % | 77 % |
| | Kappa statistics, K | **0.7** | **0.7** | 0.7 | 0.7 | 0.7 |
| Artificial neural network (ANN) | Total accuracy | 80 % | **80 %** | **80 %** | 80 % | 79 % |
| | Macro accuracy | 71 % | **72 %** | **72 %** | 71 % | 70 % |
| | Kappa statistics, K | 0.7 | **0.7** | **0.7** | 0.7 | 0.7 |
| Decision tree (DT) | Total accuracy | **76 %** | 76 % | 76 % | 76 % | 76 % |
| | Macro accuracy | **68 %** | 67 % | 66 % | 67 % | 66 % |
| | Kappa statistics, K | **0.63** | 0.63 | 0.63 | 0.63 | 0.63 |
| Support vector machine (SVM) | Total accuracy | **79 %** | 77 % | 76 % | 75 % | 75 % |
| | Macro accuracy | **68 %** | 67 % | 67 % | 67 % | 67 % |
| | Kappa statistics, K | **0.67** | 0.67 | 0.67 | 0.66 | 0.67 |
| Logistic regression (LR) | Total accuracy | **70 %** | 70 % | 69 % | 69 % | 69 % |
| | Macro accuracy | **64 %** | 63 % | 63 % | 63 % | 63 % |
| | Kappa statistics, K | **0.54** | 0.54 | 0.54 | 0.54 | 0.54 |

## Sensitivity analysis

Fig. S2, S3, and S4 in the Supplementary Materials present the results obtained from sensitivity analysis, when the algorithms were retrained based on different training sets consisting of different participant data. Overall, the predictive performance of DL and ML algorithms minimally changed when different training and test sets were used. LSTM, BiLSTM, and GRU with raw acceleration signals had an overall accuracy of approximately 83%, followed by 1D-CNN achieving a total accuracy of approximately 78%. Similarly, when retraining the algorithms with handcrafted features, the total accuracy was only 1-4% different.

## DISCUSSION

This study compared the ability of DL and classical ML algorithms for predicting 24-hour movement behavior intensities into sleep, sedentary, LPA, and MVPA from wrist-based raw accelerometer signals. Four DL algorithms were trained on raw acceleration signals and handcrafted features and were compared with six classical ML algorithms trained on handcrafted features. Our findings revealed that DL algorithms, even when trained directly with raw acceleration signals, can only slightly improve the classification accuracy of classical ML models for predicting 24-hour movement intensities [9, 44].

The best-performing DL algorithms with raw acceleration were LSTM, BiLSTM, and GRU algorithms, which were originally designed to utilize temporal and sequential information in time series data based on capturing dependencies [19]. This implies that it may be important to utilize sequential and temporal dependencies for the classification of 24-hour movement behavior intensities. Many studies using such algorithms with sequential data have observed that these DL algorithms could outperform other DL techniques [19, 33, 45]. Similar to our results, a recent study using BiLSTM to predict sitting patterns also achieved excellent accuracy [9]. Hence, these findings suggest that those DL algorithms that utilize sequential information may be better candidates for predicting 24-hour movement behaviors from raw acceleration data.

While XGBoost had superior total accuracy compared to other algorithms when utilizing handcrafted features, the RF algorithm had less confusion in predicting the LPA and MVPA categories. RF also had better classification accuracy for LPA and MVPA compared to DL methods. These findings are in line with past studies that have demonstrated that RF can perform better than other ML algorithms in predicting movement behavior intensities and types [14, 24]. Our study results extend the findings of previous studies by highlighting that RF may perform comparably to DL algorithms overall, with better classification of LPA and MVPA.

We observed that DL algorithms, when trained with handcrafted features, have slightly lower performance than when utilizing raw signals. Extracting handcrafted features from window intervals reduces the size of the original raw signals; this size reduction varies depending on the length of the segmented windows used for feature extraction. Large volumes of data enable DL algorithms to better capture a broader range of patterns and representations [46]. A potential explanation for the decrease in performance of DL algorithms when utilizing handcrafted features as opposed to raw signals could be partly explained by the reduction in data volume within the handcrafted features. This underscores the importance of having adequate training data when training DL algorithms to predict movement behaviors [45].

The confusion between MVPA and LPA was observed to be relatively higher. The resemblance between MVPA and LPA acceleration data can make their classification a challenging task [47]. Compared to both classical ML algorithms and DL algorithms with handcrafted features, DL algorithms such as LSTM, BiLSTM, and GRU exhibited reduced confusion between these two movement intensities when utilizing raw signals. Overall, LPA classification was improved by approximately 10%. This suggests that raw signals might be more effective in detecting these differences compared to handcrafted features. Such observations further highlight the potential of utilizing raw signals with more complex DL algorithms to better distinguish between these movements' classes, contributing to improved classification accuracy in 24-hour movement behaviors [8, 9].

**Strengths, limitations, and future directions**

This study has several strengths. We utilized a large, open-access dataset with 151 participants, which can potentially enhance the reproducibility of our results. The use of free-living acceleration data for model development and testing is another strength given that many studies have trained and validated ML approaches using a small dataset with limited activity types collected under laboratory conditions [16, 18, 48]. To minimize concerns of overfitting, we retrained our models while randomly relocating participants in different training and testing sets. We also retrained the algorithms using top-selected features through a feature selection algorithm, ensuring that the study's results would remain consistent with a different set of input features.

The participants in the Capture24 dataset had a wide age range, which may be considered a limitation when applying the models to participants with a limited age range. Due to the absence of measured energy expenditure, the same MET thresholds were used to categorize movement behavior intensities for all participants; this means that activities of similar types were grouped into the same intensity category, disregarding the individual variability in energy expenditure among the participants. Higher computational time and model complexity are widely recognized as inherent challenges of ML and DL algorithms. However, the models developed and compared in this study are primarily used to characterize physical behaviors in observational studies [16], where they are applied post-data collection rather than in real time. In such applications, as opposed to real-time use, training time and computational cost are generally less important. Following the literature [24], the size of the window may have an impact on the performance of ML classification models for predicting movement behaviors. We used a 10-second window, which is considered short enough to possibly also include sporadic activities in adult populations [9, 24]. Nevertheless, future studies may consider the influence of window size, particularly on the performance of DL algorithms.

## CONCLUSION

The DL algorithms trained with raw acceleration signals demonstrated slightly higher accuracy in predicting 24-hour movement behaviors into sleep, sedentary, LPA, and MVPA compared to when these algorithms were trained with handcrafted features or when classical ML algorithms were with handcrafted features. These findings underscore that DL algorithms trained directly on raw acceleration signals may be *marginally* better than classical ML that require handcrafted feature engineering. Among the DL and ML algorithms examined in this study, RF exhibited relatively less confusion between LPA and MVPA, while having a reasonably high accuracy overall. This suggests that RF could be a reasonable choice for predicting movement behavior intensities into sleep, sedentary, LPA, and MVPA.

## Declarations

**Ethics approval and consent to participate**

Not Applicable

**Consent for publication**

Not Applicable

**Availability of data and materials**

The data sets used and/or analyzed during the current study are publicly available as Capture-24 Data set.

**Competing interests**

The authors declare that they have no competing interests.

**Funding**

VF is supported by the Institute for Sport and Sport Science, TU Dortmund University, Germany. The project is supported by the Academy of Finland (project numbers 326291) and the University of Oulu. This study has also received funding from the Ministry of Education and Culture in Finland [grant numbers OKM/20/626/2022, OKM/76/626/2022, OKM/68/626/2023].

**Authors' contributions**

AS: Method, Implementation, Writing and Review

MR: Method and Review

MO: Method and Review

VF: Method, Writing, Review and Supervisory

**Acknowledgements**

**Conflict of interest**

The authors declare that there is no conflict of interest regarding the publication of this article.

# REFERENCES

1. Rosenberger ME, Fulton JE, Buman MP, Troiano RP, Grandner MA, Buchner DM, et al. The 24-Hour Activity Cycle: A New Paradigm for Physical Activity. Med Sci Sports Exerc. 2019;51:454–64.

2. Narayanan A, Desai F, Stewart T, Duncan S, MacKay L. Application of Raw Accelerometer Data and Machine-Learning Techniques to Characterize Human Movement Behavior: A Systematic Scoping Review. J Phys Act Heal. 2020;17:360–83.

3. Trost SG, Brookes DSK, Ahmadi MN. Evaluation of Wrist Accelerometer Cut-Points for Classifying Physical Activity Intensity in Youth. Front Digit Heal. 2022;4.

4. Chowdhury AK, Tjondronegoro D, Chandran V, Trost SG. Ensemble Methods for Classification of Physical Activities from Wrist Accelerometry. Med Sci Sports Exerc. 2017;49:1965–73.

5. Thapa-Chhetry B, Arguello DJ, John D, Intille S. Detecting Sleep and Nonwear in 24-h Wrist Accelerometer Data from the National Health and Nutrition Examination Survey. Med Sci Sports Exerc. 2022;54:1936–46.

6. Lambiase MJ, Gabriel KP, Chang YF, Kuller LH, Matthews KA. Utility of actiwatch sleep monitor to assess waking movement behavior in older women. Med Sci Sports Exerc. 2014;46:2301–7.

7. Fairclough SJ, Clifford L, Brown D, Tyler R. Characteristics of 24-hour movement behaviours and their associations with mental health in children and adolescents. J Act Sedentary Sleep Behav. 2023;2.

8. Farrahi V, Muhammad U, Rostami M, Oussalah M. AccNet24: A deep learning framework for classifying 24-hour activity behaviours from wrist-worn accelerometer data under free-living environments. Int J Med Inform. 2023;172.

9. Carlson JA, Ridgers ND, Nakandala S, Zablocki R, Tuz-Zahra F, Bellettiere J, et al. CHAP-child: an open source method for estimating sit-to-stand transitions and sedentary bout patterns from hip accelerometers among children. Int J Behav Nutr Phys Act. 2022;19.

10. Kerr J, Patterson RE, Ellis K, Godbole S, Johnson E, Lanckriet G, et al. Objective Assessment of Physical Activity: Classifiers for Public Health. Med Sci Sports Exerc. 2016;48:951–7.

11. Montoye AHK, Westgate BS, Fonley MR, Pfeiffer KA. Cross-validation and out-of-sample testing of physical activity intensity predictions with a wrist-worn accelerometer. J Appl Physiol. 2018;124:1284–93.

12. Farrahi V, Niemela M, Tjurin P, Kangas M, Korpelainen R, Jamsa T. Evaluating and Enhancing the Generalization Performance of Machine Learning Models for Physical Activity Intensity Prediction from Raw Acceleration Data. IEEE J Biomed Heal Informatics. 2020;24:27–38.

13. Zhang S, Rowlands A V., Murray P, Hurst TL. Physical activity classification using the GENEA wrist-worn accelerometer. Med Sci Sports Exerc. 2012;44:742–8.

14. Chong J, Tjurin P, Niemelä M, Jämsä T, Farrahi V. Machine-learning models for activity class prediction: A comparative study of feature selection and classification algorithms. Gait Posture. 2021;89:45–53.

15. Farrahi V, Rostami M. Machine learning in physical activity, sedentary, and sleep behavior research. J Act Sedentary Sleep Behav 2024 31. 2024;3:1–17.

16. Farrahi V, Niemelä M, Kangas M, Korpelainen R, Jämsä T. Calibration and validation of accelerometer-based activity monitors: A systematic review of machine-learning approaches. Gait Posture. 2019;68:285–99.

17. de Almeida Mendes M, da Silva ICM, Ramires V V., Reichert FF, Martins RC, Tomasi E. Calibration of raw accelerometer data to measure physical activity: A systematic review. Gait Posture. 2018;61:98–110.

18. Clark CCT, Barnes CM, Stratton G, McNarry MA, Mackintosh KA, Summers HD. A Review of Emerging Analytical Techniques for Objective Physical Activity Measurement in Humans. Sports Med. 2017;47:439–47.

19. Sarker IH. Deep Learning: A Comprehensive Overview on Techniques, Taxonomy, Applications and Research Directions. SN Comput Sci. 2021;2:1–20.

20. Parker PRL, Brown MA, Smear MC, Niell CM. Movement-Related Signals in Sensory Areas: Roles in Natural Behavior. Trends Neurosci. 2020;43:581–95.

21. Ao SI, Fayek H. Continual Deep Learning for Time Series Modeling. Sensors (Basel). 2023;23.

22. Greenwood-Hickman MA, Nakandala S, Jankowska MM, Rosenberg DE, Tuz-Zahra F, Bellettiere J, et al. The CNN Hip Accelerometer Posture (CHAP) Method for Classifying Sitting Patterns from Hip Accelerometers: A Validation Study. Med Sci Sports Exerc. 2021;53:2445–54.

23. Willetts M, Hollowell S, Aslett L, Holmes C, Doherty A. Statistical machine learning of sleep and physical activity phenotypes from sensor data in 96,220 UK Biobank participants. Sci Reports 2018 81. 2018;8:1–10.

24. Bach K, Kongsvold A, Bårdstu H, Bardal EM, Kjærnli HS, Herland S, et al. A Machine Learning Classifier for Detection of Physical Activity Types and Postures During Free-Living. J Meas Phys Behav. 2021;5:24–31.

25. Zhao Z, Anand R, Wang M. Maximum relevance and minimum redundancy feature selection methods for a marketing machine learning platform. Proc - 2019 IEEE Int Conf Data Sci Adv Anal DSAA 2019. 2019;:442–52.

26. Walmsley R, Chan S, Smith-Byrne K, Ramakrishnan R, Woodward M, Rahimi K, et al. Reallocation of time between device-measured movement behaviours and risk of incident cardiovascular disease. Br J Sports Med. 2021;56:1008–17.

27. Ainsworth BE, Haskell WL, Herrmann SD, Meckes N, Bassett DR, Tudor-Locke C, et al. 2011 Compendium of Physical Activities: a second update of codes and MET values. Med Sci Sports Exerc. 2011;43:1575–81.

28. Ellis K, Godbole S, Kerr J, Lanckriet G. Multi-sensor physical activity recognition in free-living. Proc . ACM Int Conf Ubiquitous Comput  UbiComp. 2014;2014:431–9.

29. Hochreiter S, Schmidhuber J. Long Short-Term Memory. Neural Comput. 1997;9:1735–80.

30. Graves A, Schmidhuber J. Framewise phoneme classification with bidirectional LSTM and other neural network architectures. Neural Netw. 2005;18:602–10.

31. Chung J, Gulcehre C, Cho K, Bengio Y. Empirical Evaluation of Gated Recurrent Neural Networks on Sequence Modeling. 2014.

32. Albawi S, Mohammed TA, Al-Zawi S. Understanding of a convolutional neural network. Proc 2017 Int Conf Eng Technol ICET 2017. 2018;2018-January:1–6.

33. Zargoush M, Sameh A, Javadi M, Shabani S, Ghazalbash S, Perri D. The impact of recency and adequacy of historical information on sepsis predictions using machine learning. Sci Rep. 2021;11.

34. Huang S, Tang J, Dai J, Wang Y. Signal Status Recognition Based on 1DCNN and Its Feature Extraction Mechanism Analysis. Sensors (Basel). 2019;19.

35. Mitchell R, Frank E. Accelerating the XGBoost algorithm using GPU computing. PeerJ Comput Sci. 2017;3:e127.

36. Breiman L. Random forests. Mach Learn. 2001;45:5–32.

37. Wang Z, Di Massimo C, Tham MT, Julian Morris A. A procedure for determining the topology of multilayer feedforward neural networks. Neural Networks. 1994;7:291–300.

38. Cortes C, Vapnik V, Saitta L. Support-vector networks. Mach Learn 1995 203. 1995;20:273–97.

39. Hastie T, Tibshirani R, Friedman J. The Elements of Statistical Learning. 2009. https://doi.org/10.1007/978-0-387-84858-7.

40. Hosmer DW, Lemeshow S, Sturdivant RX. Applied Logistic Regression: Third Edition. Appl Logist Regres Third Ed. 2013;:1–510.

41. Dalton A, Ólaighin G. Comparing supervised learning techniques on the task of physical activity recognition. IEEE J Biomed Heal informatics. 2013;17:46–52.

42. Staudenmayer J, Zhu W, Catellier DJ. Statistical considerations in the analysis of accelerometry-based activity monitor data. Med Sci Sports Exerc. 2012;44 SUPPL:S61–7.

43. [PDF] From Bias and Prevalence to Macro F1, Kappa, and MCC: A structured overview of metrics for multi-class evaluation | Semantic Scholar. https://www.semanticscholar.org/paper/From-Bias-and-Prevalence-to-Macro-F1%2C-Kappa%2C-and-A-Opitz/ff334e8cee3550eb5c80a61213ad0aecb549f48f. Accessed 11 Nov 2024.

44. Saez Y, Baldominos A, Isasi P. A Comparison Study of Classifier Algorithms for Cross-Person Physical Activity Recognition. Sensors (Basel). 2016;17.

45. Zhang S, Li Y, Zhang S, Shahabi F, Xia S, Deng Y, et al. Deep Learning in Human Activity Recognition with Wearable Sensors: A Review on Advances. Sensors (Basel). 2022;22.

46. Thakur D, Biswas S. Feature fusion using deep learning for smartphone based human activity recognition. Int J Inf Technol an Off J Bharati Vidyapeeth's Inst Comput Appl Manag. 2021;13:1615–24.

47. Ahmadi MN, Trost SG. Device-based measurement of physical activity in pre-schoolers: Comparison of machine learning and cut point methods. PLoS One. 2022;17:e0266970.

48. Narayanan A, Desai F, Stewart T, Duncan S, MacKay L. Application of Raw Accelerometer Data and Machine-Learning Techniques to Characterize Human Movement Behavior: A Systematic Scoping Review. J Phys Act Heal. 2020;17:360–83.

**List of Abbreviations:**

ML : Machine Learning

DL: Deep Learning

LSTM: Long Short Term Memory

BiLSTM: Bidirectional Long Short Term Memory

GRU: Gated Recurrent Units

1D-CNN: 1Dimentional Convoulutional Neaural Network

RF: Random Forest

XGBoost: eXtreme Gradian Boosting

DT: Decision Tree

ANN: Artificial Neaural Network

SVM: Support Vector Machine

LR: Logistic Regression

LPA: Light Physical Activity

MVPA: Moderate to Vigouros Physical Activity

# Supplementary materials File1

**Mutual score**

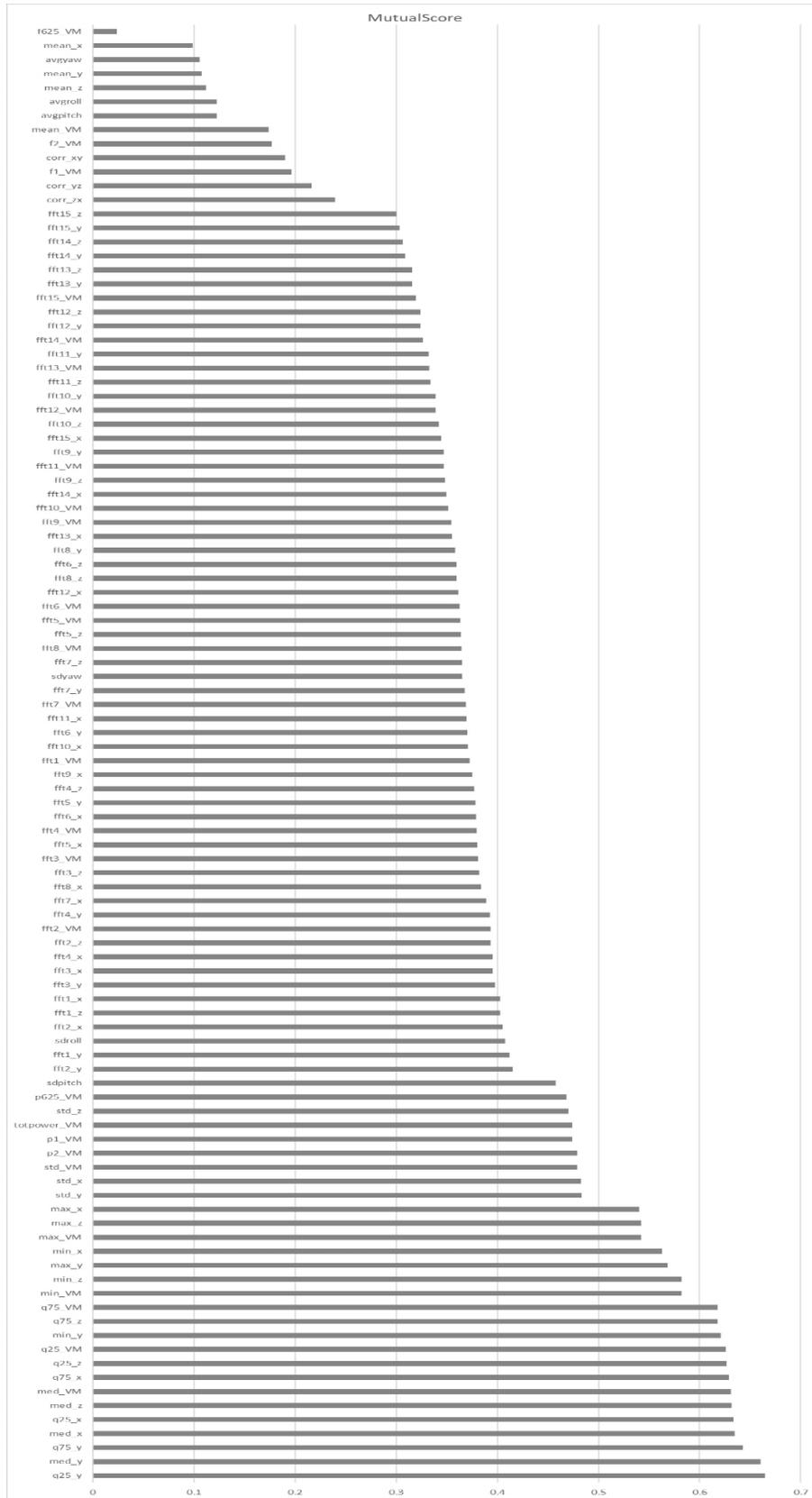

*Fig. S6 : Mutual score for features*

**Sensitivity analysis**

- DL algorithms with raw signals

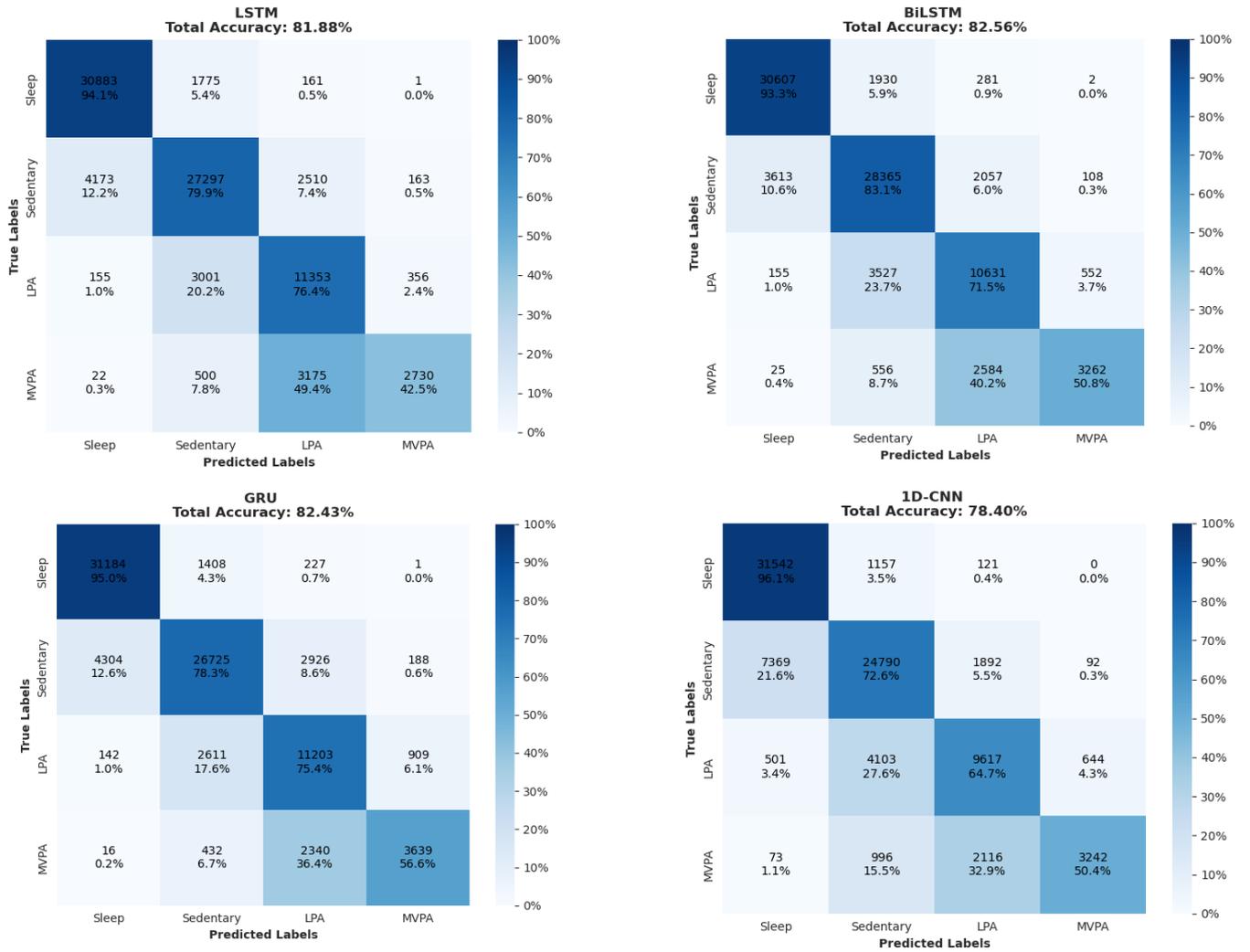

*Fig. S7: DL with Raw signals*

- DL algorithms with handcrafted features

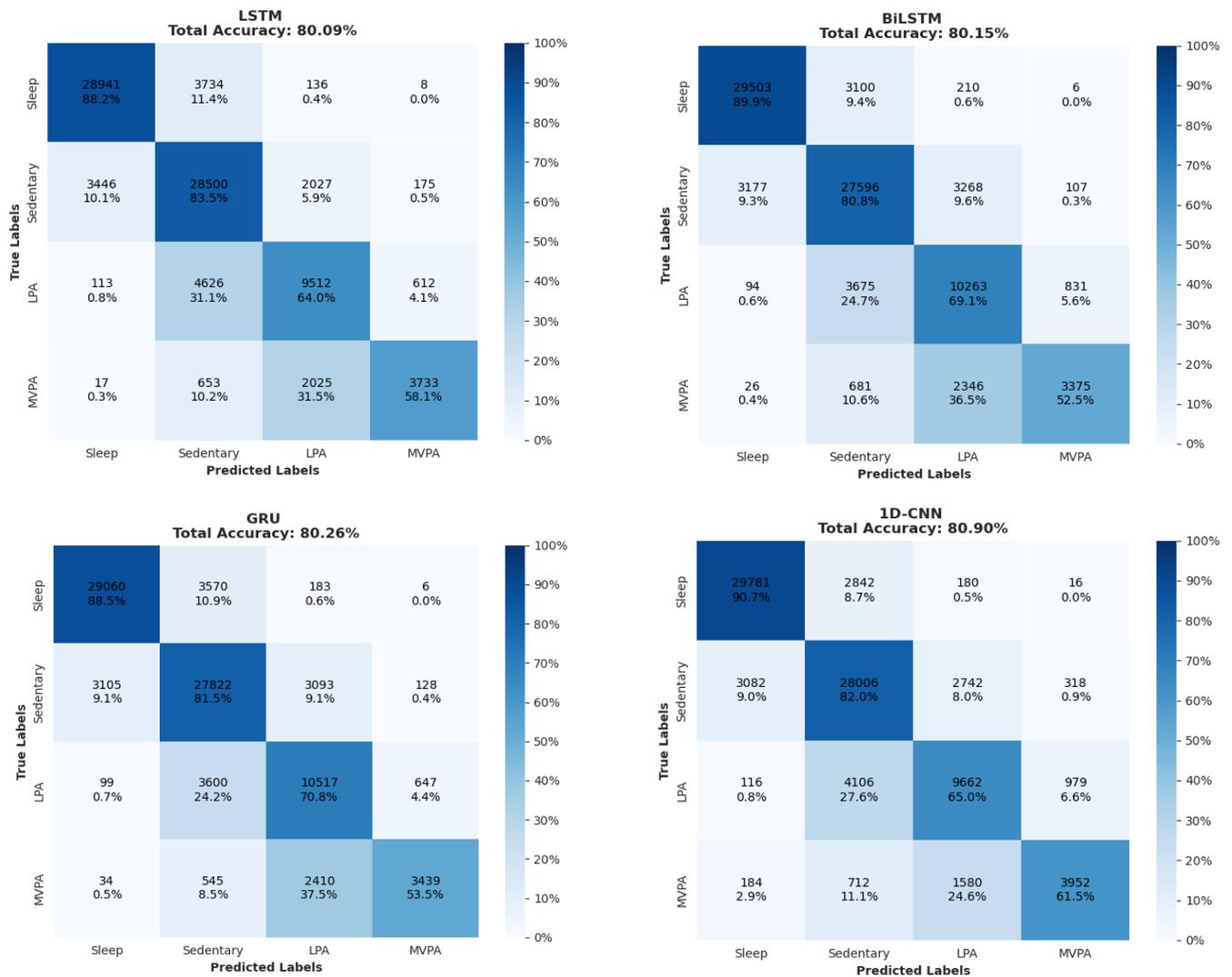

**Fig. S8: DL with handcrafted features**

- Classical ML algorithms with hand-crafted features

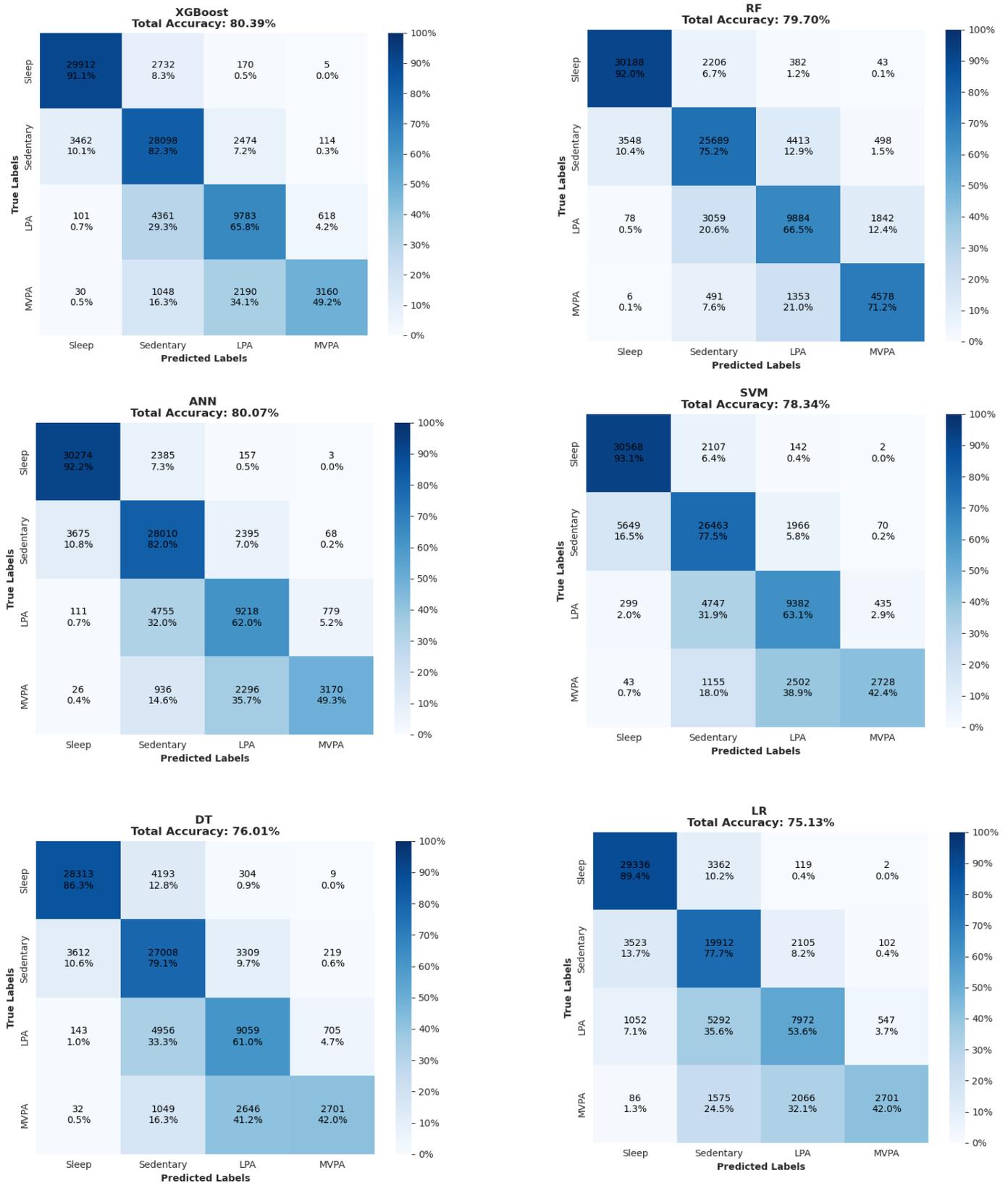

*Fig. S9: Classical ML with handcrafted features*

**Sensitivity analysis -MRMR feature selection**

- DL algorithms with hand-crafted features

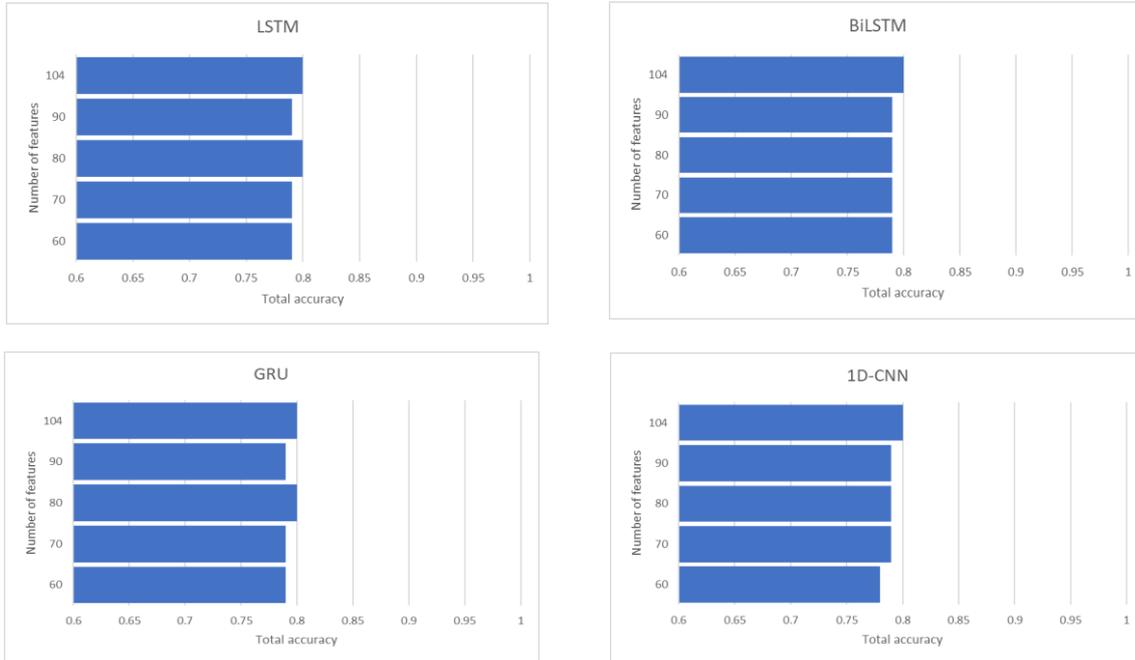

*Fig. S10: MRMR features selection with DL*

- Classical ML algorithms with hand-crafted features

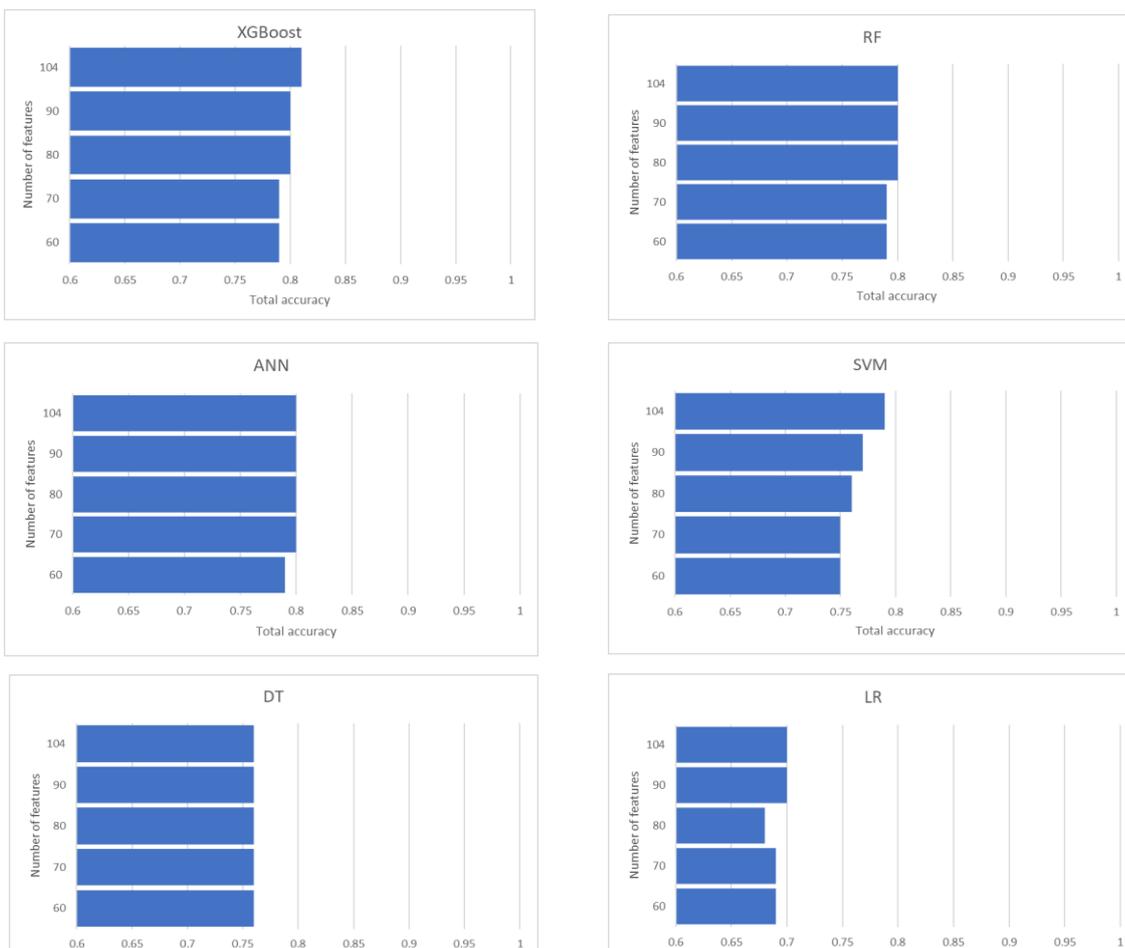

*Fig. S11: MRMR feature selection with classical ML*